\pdfoutput=1

\documentclass[11pt]{article}

\usepackage{naacl2021}

\usepackage{times}
\usepackage{latexsym}

\usepackage{bm,amsmath,amssymb,mathrsfs,amsfonts}
\usepackage{caption}
\usepackage{scalefnt}
\usepackage{multirow}
\usepackage{here}
\usepackage{booktabs}
\usepackage{url}
\usepackage{enumitem}
\usepackage{type1cm}
\usepackage{cite}
\usepackage{arydshln}
\usepackage{autobreak}
\usepackage{breqn}

\newcommand{\speech}{X}
\newcommand{\ysrc}{Y^{\rm s}}
\newcommand{\ysrchyp}{\hat{Y}^{\rm s}}
\newcommand{\ytgt}{Y^{\rm t}}
\newcommand{\ytgthyp}{\hat{Y}^{\rm t}}
\newcommand{\asrpair}{{\cal D}_{\rm asr}}
\newcommand{\fwdseqkdstpair}{{\cal D}^{\rm fwd}_{\rm st}}
\newcommand{\bwdseqkdstpair}{{\cal D}^{\rm bwd}_{\rm st}}
\newcommand{\bidirseqkdstpair}{{\cal D}^{\rm bidir}_{\rm st}}
\newcommand{\stpair}{{\cal D}_{\rm st}}
\newcommand{\mtpair}{{\cal D}_{\rm mt}}

\newcommand{\fwdmt}{{\cal M}_{\rm fwd}}
\newcommand{\bwdmt}{{\cal M}_{\rm bwd}}

\usepackage[T1]{fontenc}

\usepackage[utf8]{inputenc}

\usepackage{microtype}

\title{Source and Target Bidirectional Knowledge Distillation\\for End-to-end Speech Translation}

\author{Hirofumi Inaguma${}^{\clubsuit}$ ~ Tatsuya Kawahara$^{\clubsuit}$ ~ \textbf{Shinji Watanabe}${}^{\spadesuit}$ \\
${}^{\clubsuit}$ Kyoto University, Japan ~~ ${}^{\spadesuit}$ Johns Hopkins University, USA\\
\texttt{inaguma@sap.ist.i.kyoto-u.ac.jp}
}

\begin{document}
\maketitle
\begin{abstract}
A conventional approach to improving the performance of end-to-end speech translation (E2E-ST) models is to leverage the source transcription via pre-training and joint training with automatic speech recognition (ASR) and neural machine translation (NMT) tasks.
However, since the input modalities are different, it is difficult to leverage source language text successfully.
In this work, we focus on sequence-level knowledge distillation (SeqKD) from external text-based NMT models.
To leverage the full potential of the source language information, we propose \textit{backward SeqKD}, SeqKD from a target-to-source backward NMT model.
To this end, we train a bilingual E2E-ST model to predict paraphrased transcriptions as an auxiliary task with a single decoder.
The paraphrases are generated from the translations in bitext via back-translation.
We further propose \textit{bidirectional SeqKD} in which SeqKD from both forward and backward NMT models is combined.
Experimental evaluations on both autoregressive and non-autoregressive models show that SeqKD in each direction consistently improves the translation performance, and the effectiveness is complementary regardless of the model capacity.
\end{abstract}

\section{Introduction}
End-to-end speech translation (E2E-ST)~\citep{listen_and_translate}, which aims to convert source speech to text in another language directly, is an active research area.
Because direct ST is a more difficult task than automatic speech recognition (ASR) and machine translation (MT), various techniques have been proposed to ease the training process by using source transcription.
Examples include pre-training~\citep{berard2018end,wang2020bridging,bansal-etal-2019-pre,wang-etal-2020-curriculum}, multi-task learning~\citep{weiss2017sequence,berard2018end,bahar2019comparative}, knowledge distillation~\citep{liu2019end}, meta-learning~\citep{indurthi2020end}, two-pass decoding~\citep{anastasopoulos-chiang-2018-tied,sperber-etal-2019-attention}, and interactive decoding~\citep{liu2020synchronous,le-etal-2020-dual}.
However, as input modalities between ST and MT tasks are different, an auxiliary MT task is not always helpful, especially when additional bitext is not available~\citep{bahar2019comparative}.
Moreover, because monotonic speech-to-transcription alignments encourage the ASR task to see surface-level local information,
an auxiliary ASR task helps the E2E-ST model to extract acoustic representations, not semantic ones, from speech.

Sequence-level knowledge distillation (SeqKD)~\citep{kim-rush-2016-sequence} is another approach to transferring knowledge from one model to another.
Recent studies have shown that SeqKD has the effect of reducing the complexity of training data and thus eases the training of student models, e.g., non-autoregressive (NAR) models~\citep{gu2018non,zhou2019understanding,ren-etal-2020-study}.

Paraphrasing, which represents text in a different form but with the same meaning, can also be regarded as SeqKD when using neural paraphrasing via back-translation~\citep{mallinson-etal-2017-paraphrasing,wieting-etal-2017-learning,federmann-etal-2019-multilingual}.
It has been studied to improve the reference diversity for MT system evaluations~\citep{thompson-post-2020-automatic,bawden2020parbleu,bawden-etal-2020-study} and the performance of low-resource neural MT (NMT) models~\citep{zhou-etal-2019-paraphrases,khayrallah-etal-2020-simulated}. 

In this work, due to its simplicity and effectiveness, we focus on SeqKD from text-based NMT models to improve the performance of a bilingual E2E-ST model.
In order to fully leverage source language information, we propose \textit{backward SeqKD}, which targets paraphrased source transcriptions generated from a target-to-source backward NMT model as an auxiliary task.
Then, a single ST decoder is trained to predict both source and target language text as in a multilingual setting~\citep{inaguma19asru}.
This way, the decoder is biased to capture semantic representations from speech, unlike joint training with an auxiliary ASR task.
We also propose \textit{bidirectional SeqKD}, which combines SeqKD from two NMT models in both language directions.
Therefore, the E2E-ST models can fully exploit the knowledge embedded in both forward and backward NMT models.

Experimental evaluations demonstrate that SeqKD from each direction consistently improves the translation performance of both autoregressive and non-autoregressive E2E-ST models.
We also confirm that bidirectional SeqKD outperforms unidirectional SeqKD and that the effectiveness is maintained in large models.

\section{Method}\label{sec:proposed_method}
In this section, we propose bidirectional SeqKD from both forward and backward NMT models that leverages machine-generated source paraphrases as another target in addition to the distilled translation to enhance the training of a bilingual E2E-ST model.
Let $\speech$ denote input speech features in a source language and $\ysrc$ and $\ytgt$ denote the corresponding gold transcription and translation, respectively.
Let $\stpair=\{(\speech_{i}, \ysrc_{i}, \ytgt_{i})\}_{i=1}^{I}$ be an ST dataset including $I$ samples, and $\asrpair=\{(\speech_{i}, \ysrc_{i})\}_{i=1}^{I}$ and $\mtpair=\{(\ysrc_{i}, \ytgt_{i})\}_{i=1}^{I}$ denote the corresponding ASR and MT datasets, respectively.\footnote{We focus on a complete triplet of ($\speech$, $\ysrc$, $\ytgt$) only. However, the proposed method can easily be extended to a semi-supervised setting featuring additional ASR and MT pair data.}
We drop the subscript $i$ when it is obvious.

\subsection{Sequence-level knowledge distillation}\label{ssec:seq_kd}
We first train a text-based source-to-target forward NMT model $\fwdmt$ with $\mtpair$.\footnote{All NMT models are autoregressive in this paper.}
Then, we perform beam search decoding with $\fwdmt$ on $\stpair$ to create a new dataset $\fwdseqkdstpair=\{(\speech_{i}, \ysrc_{i}, \ytgthyp_{i})\}_{i=1}^{I}$, where $\ytgthyp_{i}$ is a distilled translation.
$\fwdseqkdstpair$ is used to train the E2E-ST models, referred to as \textit{forward SeqKD} (or fwd SeqKD).

\subsection{Paraphrase generation}\label{ssec:paraphrase_generation}
To exploit semantic information in the source language, we leverage machine-generated paraphrases of source transcriptions.
We train a text-based target-to-source backward NMT model $\bwdmt$ with $\mtpair$ and then generate a new dataset $\bwdseqkdstpair=\{(\speech_{i}, \ysrchyp_{i}, \ytgt_{i})\}_{i=1}^{I}$, where $\ysrchyp_{i}$ is a paraphrase of $\ysrc_{i}$.
We use $\bwdseqkdstpair$ for training the E2E-ST models.
As neural paraphrasing can be regarded as SeqKD from $\bwdmt$, we referred to it as \textit{backward SeqKD} (or bwd SeqKD).
In this work, we do not use large paraphrase datasets~\citep{wieting-gimpel-2018-paranmt,hu2019parabank} because their availability depends on languages and domains.
Moreover, neural paraphrasing is applicable to any source languages that lack a sufficient amount of paired paraphrase data.

We also propose combining forward SeqKD with backward SeqKD, referred to as \textit{bidirectional SeqKD} (or bidir SeqKD), and construct a new dataset $\bidirseqkdstpair=\{(\speech_{i}, \ysrchyp_{i}, \ytgthyp_{i})\}_{i=1}^{I}$.
When using two references per utterance (\textit{2ref} training)~\citep{gordon2019explaining}, we concatenate $\fwdseqkdstpair$ and $\bwdseqkdstpair$, and the suitable combination is analyzed in Section~\ref{ssec:ablation}.
This way, we can distill the knowledge of both $\fwdmt$ and $\bwdmt$ to a single E2E-ST model.

\subsection{Training}\label{ssec:training}
We train an E2E-ST model with a direct ST objective ${\cal L}_{\rm st}(\ytgt \ {\rm or} \ \ytgthyp|\speech)$ and an auxiliary speech-to-source text objective ${\cal L}_{\rm src}(\ysrc \ {\rm or} \ \ysrchyp|\speech)$.
We refer to joint training with ${\cal L}_{\rm src}(\ysrc|\speech)$ as \textit{joint ASR} and with ${\cal L}_{\rm src}(\ysrchyp|\speech)$ as \textit{backward SeqKD}.
Both losses are calculated from the same ST decoder.
To bias the model to generate the desired target language, we add language embedding to token embedding at \textit{every} token position in the decoder~\citep{conneau2019cross}.\footnote{We found this was more effective than replacing the start-of-sentence symbol with a language ID~\citep{inaguma19asru,wang2020covost,le-etal-2020-dual} as done in previous multilingual E2E-ST studies.}
We then apply bidirectional SeqKD to both autoregressive (AR) and non-autoregressive (NAR) E2E-ST models.

\subsubsection*{Autoregressive E2E-ST model}
We use the speech Transformer architecture in~\citep{karita2019comparative} with an additional language embedding.
The total training objective is formulated with a hyperparameter $\lambda_{\rm src} (\geq 0)$ as
\begin{eqnarray}
{\cal L}_{\rm total} = {\cal L}_{\rm st} + \lambda_{\rm src} {\cal L}_{\rm src}, \label{eq:total_loss}
\end{eqnarray}
where both ${\cal L}_{\rm st}$ and ${\cal L}_{\rm src}$ are defined as cross-entropy losses.
The entire encoder-decoder parameters are shared in both tasks.

\subsubsection*{Non-autoregressive E2E-ST model}
We adopt Orthors~\citep{inaguma2020orthros}, in which a decoder based on a conditional masked language model (CMLM)~\citep{ghazvininejad-etal-2019-mask} is jointly trained with an additional AR decoder on the shared speech encoder.
The training of the NAR decoder is further enhanced with semi-autoregressive training (SMART)~\citep{ghazvininejad2020semi}.
${\cal L}_{\rm st}$ in Eq.~\eqref{eq:total_loss} is modified as
\begin{eqnarray}
{\cal L}_{\rm st} = {\cal L}_{\rm cmlm} + \lambda_{\rm ar} {\cal L}_{\rm ar} + \lambda_{\rm lp} {\cal L}_{\rm lp}, \label{eq:st_loss_nar}
\end{eqnarray}
where ${\cal L}_{\rm cmlm}$, ${\cal L}_{\rm ar}$, and ${\cal L}_{\rm lp}$ are losses in NAR E2E-ST, AR E2E-ST, and length prediction tasks, respectively.
$\lambda_{\rm *}$ is the corresponding tunable loss weight.
During inference, the mask-predict algorithm is used for $T$ iterations with a length beam width of $l$~\citep{ghazvininejad-etal-2019-mask}.
The best candidate at the last iteration is selected from the NAR decoder based on scores from the AR decoder~\citep{inaguma2020orthros}.
Note that we apply ${\cal L}_{\rm src}$ to the NAR decoder only.

\begin{table}[t]
    \centering
    \begingroup
    \scalebox{1.00}{
    \begin{tabular}{lcc}\toprule
     Language direction & \bf{BLEU} ($\uparrow$) & \bf{TER} ($\downarrow$) \\ 
    \midrule
     De $\to$ En & 43.49 & 38.60  \\ %
     Fr $\to$ En & 48.55 & 34.30 \\ %
      \bottomrule
    \end{tabular}
    }
    \endgroup
    \vspace{-2mm}
    \caption{Quality of paraphrases in the training set}
    \label{tab:bleu_paraphrase}
    \vspace{-2mm}
\end{table}

\section{Experimental setting}\label{sec:experiment}
\paragraph{Data} 
We used Must-C En-De ($408$ hours) and En-Fr ($492$ hours) datasets~\citep{di-gangi-etal-2019-must}.
Both language pairs consist of a triplet of ($\speech$, $\ysrc$, $\ytgt$).
We performed the same data preprocessing as~\citep{inaguma-etal-2020-espnet} (see details in Appendix~\ref{appendix:sec:preprocessing}).
We report case-sensitive detokenized BLEU scores~\citep{papineni-etal-2002-bleu} on the {\tt tst-COMMON} set with the \texttt{multi-bleu-detok.perl} script in Moses~\citep{koehn-etal-2007-moses}.

\paragraph{Model configuration}
We used the Transformer~\citep{vaswani2017attention} architecture having $12$ encoder layers following two CNN blocks and six decoder layers for the ASR and E2E-ST tasks.
For the MT models, we used six encoder layers.
We built our models with the ESPnet-ST toolkit~\citep{inaguma-etal-2020-espnet}.
See details in Appendix~\ref{appendix:sec:model_configuration}.

\paragraph{Training}
We always initialized the encoder parameters of the E2E-ST model by those of the corresponding pre-trained ASR model~\citep{berard2018end}.
We follow the same optimization strategies as in~\citep{inaguma2020orthros,inaguma-etal-2020-espnet}.
When using joint ASR or backward SeqKD, we set $\lambda_{\rm src}$ to $0.3$.
More details are described in Appendix~\ref{appendix:sec:initialization} and~\ref{appendix:sec:training}.

\paragraph{Inference}
For the AR models, we used a beam width of $4$.
For the NAR models, we set $T=\{4,10\}$ and $l=9$ as in~\citep{inaguma2020orthros}.

\begin{table}[t]
    \centering
    \begingroup
    \scalebox{0.77}{
    \begin{tabular}{clcccc}\toprule
    \multicolumn{1}{c}{\multirow{2}{*}{ID}} & \multirow{2}{*}{Model} & \multicolumn{4}{c}{\bf{BLEU} ($\Delta$) ($\uparrow$)} \\ 
    \cmidrule{3-6}
    & & \multicolumn{2}{c}{En-De} & \multicolumn{2}{c}{En-Fr} \\ \midrule
      \multirow{3}{*}{--} & ESPnet-ST${}^\dagger$ & 22.91 & & 32.69 &  \\
      & Fairseq-S2T${}^\ddagger$ & 22.7\phantom{0} & & 32.9\phantom{0} &  \\ 
      & + Multilingual${}^{\diamond}$ & 24.5\phantom{0} & & 34.9\phantom{0} &  \\ 
      \midrule

      {\tt A1} & Baseline & 22.77 & & 33.51 & \\
      {\tt A2} & + MT pre-training & \bf{23.12} & (+0.35) & \bf{33.84} & (+0.33) \\
      {\tt A3} & + Joint ASR & 22.97 & (+0.20) & 33.37 & (--0.14) \\
      {\tt A4} & + Bwd SeqKD & \bf{23.11} & (+0.34) & \bf{33.78} & (+0.23) \\ 
      \cdashline{1-6}

      {\tt B1} & {\tt A1} + Fwd SeqKD & 24.42 & (+1.65) & 34.66 & (+1.15) \\
      {\tt B2} & + MT pre-training & 24.68 & (+1.91) & 34.57 & (+1.06) \\
      {\tt B3} & + Joint ASR & 24.67 & (+1.90) & 34.68 & (+1.17) \\
      {\tt B4} & + Original (2ref) & \bf{24.83} & (\bf{+2.06}) & \bf{34.92} & (\bf{+1.41}) \\
      \cdashline{1-6}
      
      {\tt C1} & {\tt A1} + Bidir SeqKD & \bf{24.83} & (\bf{+2.06}) & \bf{34.78} & (\bf{+1.27}) \\
      {\tt C2} & + Original (2ref) & \bf{25.28} & (\bf{+2.51}) & \bf{35.29} & (\bf{+1.78}) \\
      \bottomrule
    \end{tabular}
    }
    \endgroup
    \vspace{-2mm}
    \caption{BLEU scores of \underline{AR models} on Must-C \texttt{tst-COMMON} set. ${}^\dagger$~\citep{inaguma-etal-2020-espnet},
    ${}^\ddagger$~\citep{wang-etal-2020-fairseq}. ${}^{\diamond}$Large model trained with eight language pairs~\citep{wang-etal-2020-fairseq}.}\label{tab:result_mustc_ar}
    \vspace{-2mm}
\end{table}

\section{Results}\label{sec:result}
\vspace{-1.5mm}
\subsection{Main results}
We first report the paraphrasing quality, which is shown in Table~\ref{tab:bleu_paraphrase}.
As confirmed by the BLEU and translation edit rate (TER) scores~\citep{snover2006study}, the paraphrased source text was not just a simple copy of the transcription (see examples in Appendix~\ref{appendix:sec:case_study}).

\paragraph{Autoregressive models}
The results are shown in Table~\ref{tab:result_mustc_ar}.
Pre-training the ST decoder with the forward MT decoder ({\tt A2}) improved the baseline performance ({\tt A1}).
Joint ASR showed a marginal improvement on En-De but a degraded performance on En-Fr ({\tt A3}).
We attribute this to the fact that the ASR task was more trivial than the ST task and biased the shared decoder to capture surface-level textual information.
In contrast, backward SeqKD showed small but consistent improvements in both language directions ({\tt A4}), and it was as effective as MT pre-training.
As the encoder was already pre-trained with the ASR model, paraphrases had an additional positive effect on the BLEU improvement.

Forward SeqKD significantly improved the performance, as previously reported in~\citep{inaguma2020orthros}.
However, the gains by MT pre-training and joint ASR were diminished.
Forward SeqKD was more effective than backward SeqKD solely ({\tt A4} vs. {\tt B1}).
However, backward SeqKD was still beneficial on top of forward SeqKD ({\tt C1}, i.e., bidirectional SeqKD) while joint ASR was less so ({\tt B3}).
We also augmented the target translations by concatenating $\stpair$ and $\fwdseqkdstpair$ (\textit{2ref} training), which further improved forward SeqKD ({\tt B4}).
Nevertheless, a combination of \textit{2ref} training and backward SeqKD (i.e., bidirectional SeqKD with $\fwdseqkdstpair \cup \bwdseqkdstpair$) had a complementary effect and showed the best result ({\tt C2}).
It even outperformed larger multilingual models~\citep{wang-etal-2020-fairseq} without using additional data in other language pairs.

\begin{table}[t]
    \centering
    \begingroup
    \scalebox{0.86}{
    \begin{tabular}{lccc}\toprule
    \multirow{2}{*}{Model} & \multirow{2}{*}{$T$} & \multicolumn{2}{c}{\bf{BLEU} ($\uparrow$)} \\ 
    \cmidrule{3-4}
    & & En-De & En-Fr \\ \midrule
      Fwd SeqKD & \multirow{3}{*}{4} & 21.93 & 30.46 \\
      \ + Joint ASR & & 22.13 & 30.80 \\
      Bidir SeqKD & & \bf{22.22} & \bf{31.21} \\ 
      \midrule
      
      \citep{inaguma2020orthros} & \multirow{4}{*}{10} & 22.88 & 32.20 \\ 
      Fwd SeqKD (ours) &  & 22.96 & 32.42 \\
      \ + Joint ASR & & 23.31 & 32.41 \\
      Bidir SeqKD & & \bf{23.41} & \bf{32.64} \\
      \bottomrule
    \end{tabular}
    }
    \endgroup
    \vspace{-2mm}
    \caption{BLEU scores of \underline{NAR models} on Must-C \texttt{tst-COMMON} set. All methods used forward SeqKD.}\label{tab:result_mustc_nar}
    \vspace{-2mm}
\end{table}

\paragraph{Non-autoregressive models}
The results are presented in Table~\ref{tab:result_mustc_nar}.
Following the standard practice in NAR models~\citep{gu2018non}, we always used forward SeqKD.
We did not use \textit{2ref} training for the NAR models because it increases the multi-modality.
Joint ASR improved the performance on all NAR models, except for En-Fr with the number of iterations $T=10$.
However, bidirectional SeqKD with $\bidirseqkdstpair$ further improved the performance consistently regardless of $T$.
Since NAR models assume conditional independence for every token, they prefer monotonic input-output alignments with lower alignment complexity in theory.
However, paraphrasing collapses the monotonicity of the ASR task and increases the alignment complexity, making the auxiliary speech-to-source text task non-trivial.
Nevertheless, BLEU scores were improved by adding backward SeqKD.
This was probably because the complexity of transcriptions in the training data was reduced at the cost of the alignment complexity, which was more effective for the NAR models.

\begin{table}[t]
    \centering
    \begingroup
    \scalebox{0.80}{
    \begin{tabular}{lcc}\toprule
    \multirow{2}{*}{Condition} & \multicolumn{2}{c}{\shortstack{\bf{Entropy}\\($\uparrow$ more complex)}} \\ 
    \cmidrule{2-3}
       & En-De & En-Fr \\ \midrule
     ${\cal C}(\overrightarrow{\stpair})$ (Real) & 0.70 & 0.65 \\
      ${\cal C}(\overrightarrow{\fwdseqkdstpair})$ (Fwd SeqKD) & \bf{0.52} & \bf{0.47} \\ 
      ${\cal C}(\overrightarrow{\bwdseqkdstpair})$ (Bwd SeqKD) & 0.54 & 0.47 \\
      ${\cal C}(\overrightarrow{\bidirseqkdstpair})$ (Bidir SeqKD) & 0.63 & 0.61 \\ 
      \midrule
      
    ${\cal C}(\overleftarrow{\stpair})$ (Real) & 0.40 & 0.54 \\
      ${\cal C}(\overleftarrow{\fwdseqkdstpair})$ (Fwd SeqKD) & 0.28 & 0.36 \\
      ${\cal C}(\overleftarrow{\bwdseqkdstpair})$ (Bwd SeqKD) & \bf{0.25} & \bf{0.31} \\
      ${\cal C}(\overleftarrow{\bidirseqkdstpair})$ (Bidir SeqKD) & 0.37 & 0.49 \\
      \bottomrule
    \end{tabular}
    }
    \endgroup
    \vspace{-2mm}
    \caption{Corpus-level conditional entropy}
    \label{tab:analysis_complexity}
    \vspace{-2mm}
\end{table}

\begin{table}[t]
    \centering
    \begingroup
    \scalebox{0.80}{
    \begin{tabular}{lcc}\toprule
    \multirow{2}{*}{Condition} & \multicolumn{2}{c}{\shortstack{\bf{Faithfulness}\\($\downarrow$ more faithful)}} \\ 
    \cmidrule{2-3}
       & En-De & En-Fr \\ \midrule
    ${\cal F}(\overrightarrow{\fwdseqkdstpair})$ (Fwd SeqKD) & 12.61 & 11.65 \\ 
      ${\cal F}(\overrightarrow{\bwdseqkdstpair})$ (Bwd SeqKD) & \bf{\phantom{0}9.31} & \bf{\phantom{0}8.67} \\ 
      ${\cal F}(\overrightarrow{\bidirseqkdstpair})$ (Bidir SeqKD) & 11.42 & 10.72 \\ 
      \midrule

    ${\cal F}(\overleftarrow{\fwdseqkdstpair})$ (Fwd SeqKD) & \bf{\phantom{0}9.58} & \bf{\phantom{0}8.48} \\ 
      ${\cal F}(\overleftarrow{\bwdseqkdstpair})$ (Bwd SeqKD) & 12.97 & 10.70 \\ 
      ${\cal F}(\overleftarrow{\bidirseqkdstpair})$ (Bidir SeqKD) & 11.23 & \phantom{0}9.98 \\ 
      \bottomrule
    \end{tabular}
    }
    \endgroup
    \vspace{-2mm}
    \caption{Faithfulness to training data distribution}\label{tab:analysis_faithfulness}
    \vspace{-2mm}
\end{table}

\subsection{Analysis}\label{ssec:analysis}
We analyze the performance of bidirectional SeqKD through a lens of complexity in the training data following~\citep{zhou2019understanding}.
We aligned words in every source and target sentence pair with \textit{fast_align}\footnote{\url{https://github.com/clab/fast_align}}~\citep{dyer-etal-2013-simple}.
Then, we calculated corpus-level conditional entropy ${\cal C}({\cal D})$ and faithfulness ${\cal F}({\cal D})$ for both forward ($\overrightarrow{{\cal D}}$) and backward ($\overleftarrow{{\cal D}}$) language directions to evaluate the multi-modality.
In short, conditional entropy measures uncertainty of translation, and faithfulness is defined as Kullback–Leibler divergence and measures how close the distilled data distribution is to the real data distribution.
See the mathematical definition in Appendix~\ref{appendix:mathematical_formulation}.

The results of entropy and faithfulness are shown in Tables~\ref{tab:analysis_complexity} and ~\ref{tab:analysis_faithfulness}, respectively.
Consistent with~\citep{zhou2019understanding}, the entropy of target translations was reduced by forward SeqKD, indicating target translations were converted into a more deterministic and simplified form.
Interestingly, the entropy of the original translations was also reduced by backward SeqKD.
In other words, backward SeqKD modified transcriptions so that the target translations can be predicted easier.
This would help E2E-ST models learn relationships between source and target languages \textit{from speech} because E2E-ST models are not conditioned on text in another language explicitly.
Therefore, we presume that the encoder representations were enhanced by backward SeqKD.
Using machine-generated sequences in both languages increased the entropy, probably due to error accumulation.
However, E2E-ST models do not suffer from it because they are conditioned on the source speech.
We also confirmed similar trends in the reverse language direction.

Regarding faithfulness, distilled target sequences degraded faithfulness as expected.
However, an interesting finding was that the faithfulness of bidirectional SeqKD was better than that of forward SeqKD, meaning that the former reflected the true word alignment distribution more faithfully than the latter.
Although lexical choice might be degraded by targeting distilled text in both languages~\citep{ding2021understanding}, mixing the original and distilled text by \textit{2ref} training would recover it.

\begin{table}[t]
    \centering
    \begingroup
    \scalebox{0.67}{
    \begin{tabular}{lcccc}\toprule
    \multirow{2}{*}{Training data} & \multirow{2}{*}{Target1} & \multirow{2}{*}{Target2} & \multicolumn{2}{c}{\bf{BLEU} ($\uparrow$)} \\ 
    \cmidrule{4-5}
       & & & En-De & En-Fr \\ \midrule
      $\stpair \cup \fwdseqkdstpair$ ({\tt B4} + Joint ASR) & $(\bm{\ysrc}, \ytgt)$ & $(\bm{\ysrc}, \ytgthyp)$ & 25.00 & 35.05 \\
      $\stpair \cup \bidirseqkdstpair$ & $(\bm{\ysrc}, \ytgt)$ & $(\bm{\ysrchyp}, \ytgthyp)$ & 25.21 & 35.17 \\
      $\bwdseqkdstpair \cup \bidirseqkdstpair$ & $(\bm{\ysrchyp}, \ytgt)$ & $(\bm{\ysrchyp}, \ytgthyp)$ & 25.01 & 35.22 \\
      $\fwdseqkdstpair \cup \bwdseqkdstpair$ ({\tt C2}) & $(\bm{\ysrchyp}, \ytgt)$ & $(\bm{\ysrc}, \ytgthyp)$ & \bf{25.28} & \bf{35.29} \\
      \bottomrule
    \end{tabular}
    }
    \endgroup
    \vspace{-2mm}
    \caption{Ablation study of dataset concatenation on Must-C \texttt{tst-COMMON} set. \textit{2ref} training was used.}
    \label{tab:ablation_study}
    \vspace{-2mm}
\end{table}

\subsection{Ablation study}\label{ssec:ablation}
We conduct an ablation study to verify the analysis in the previous section.
In Table~\ref{tab:analysis_complexity}, we observed that it was better to have the original reference in the target sequence of either the source or target language.
For example, to reduce the entropy of German text in the training set, it was best to condition the distilled German translation on the original English transcription, and vice versa.
Therefore, we hypothesize that the best way to reduce the entropy in both source and target languages during \textit{2ref} training is to combine ($\ysrchyp$, $\ytgt$) and ($\ysrc$, $\ytgthyp$) for each sample.
We compared four ways to leverage source text: gold transcription $\ysrc$ only, distilled paraphrase $\ysrchyp$ only, and both.\footnote{Both gold translation $\ytgt$ and distilled translation $\ytgthyp$ were always used as target sequences.}
The results are shown in Table~\ref{tab:ablation_study}.
We confirmed that the model trained with the original reference in either language for every target achieved the best BLEU score, which verifies our hypothesis.

\begin{table}[t]
    \centering
    \begingroup
    \scalebox{0.90}{
    \begin{tabular}{lcc}\toprule
    \multirow{2}{*}{Model} & \multicolumn{2}{c}{\bf{BLEU} ($\uparrow$)} \\ 
    \cmidrule{2-3}
       & En-De & En-Fr \\ \midrule
      Transformer Large + Fwd SeqKD & 25.19 & 35.47 \\
      \ + Bidir SeqKD & \bf{25.62} & \bf{35.74} \\ \cdashline{1-3}
      
      Conformer + Fwd SeqKD & 26.81 & 37.23 \\
      \ + Bidir SeqKD & \bf{27.01} & \bf{37.33} \\ \hline

      Text-based NMT (WER: 0\%)$\dagger$ & 27.56 & 39.09 \\
      \bottomrule
    \end{tabular}
    }
    \endgroup
    \vspace{-2mm}
    \caption{BLEU scores of \underline{large AR models} on Must-C \texttt{tst-COMMON} set. \textit{2ref} training was used. $\dagger$ Punctuation and case information is removed on the source side.}
    \label{tab:large_model}
    \vspace{-2mm}
\end{table}

\subsection{Increasing model capacity}
Finally, we investigate the effectiveness of bidirectional Seq-KD with \textit{2ref} training when increasing the model capacity in Table~\ref{tab:large_model}.
The purpose of this experiment is to verify our expectation that large models can model complex target distributions in multi-referenced training better.
In addition to simply increasing the model dimensions, we also investigate Conformer~\citep{gulati2020}, a Transformer encoder augmented by a convolution module.
We confirmed that bidirectional SeqKD always outperformed forward SeqKD in both language directions regardless of model configurations.
We also found that the Conformer encoder significantly boosted the translation performance of forward SeqKD, but the gains of bidirectional SeqKD were transferred.

\section{Conclusion}\label{sec:conclusion}
To fully leverage knowledge in both source and target language directions for bilingual E2E-ST models, we have proposed bidirectional SeqKD, in which 
both forward SeqKD from a source-to-target NMT model and backward SeqKD from a target-to-source NMT model are combined.
Backward SeqKD is performed by targeting source paraphrases generated via back-translation from the original translations in bitext.
Then, the E2E-ST model is enhanced by training to generate both source and target language text with a single decoder.
We experimentally confirmed that SeqKD from each direction boosted the translation performance of both autoregressive and non-autoregressive E2E-ST models, and the effectiveness was additive.
Multi-referenced training with the original and distilled text gave further gains.
We also showed that bidirectional SeqKD was effective regardless of model sizes.

\section*{Acknowledgement}
The authors thank the anonymous reviewers for useful suggestions and Siddharth Dalmia, Brian Yan, and Pengcheng Guo for helpful discussions.

\bibliography{reference}
\bibliographystyle{acl_natbib}

\begin{table*}[t]
    \centering
    \small
    \begingroup
    \begin{tabular}{l|l}\toprule
      Reference1 & She took our order, and then went to the couple in the booth next to us, and she \\
      & lowered her voice so much, I had to really strain to hear what she was saying. \\ \cdashline{1-2}
      Paraphrase1 (Backward NMT) & She picked up our order, and then went to the pair in the niche next to us and \\ 
      & lowered her voice so much that I had to really try to understand them. \\ \hline
            
      Reference2 & And she said "Yes, that's former Vice President Al Gore and his wife, Tipper." \\
      & And the man said, "He's come down a long way, hasn't he?" (Laughter) \\ \cdashline{1-2}
      Paraphrase2 (Backward NMT) & She said, "Yes, that's ex-vice President Al Gore and his wife Tipper." And the \\
      & man said, "It's a nice gap, what?" (Laughter) \\
      \bottomrule
    \end{tabular}
    \endgroup
    \vspace{-2mm}
    \caption{Examples of source paraphrases on the Must-C En-De training set}\label{tab:result_case_study}
    \vspace{-2mm}
\end{table*}

\newpage
\appendix
\section{Appendix}\label{sec:appendix}
\subsection{Data preprocessing}\label{appendix:sec:preprocessing}
All sentences were tokenized with the {\tt tokenizer.perl} script in Moses~\citep{koehn-etal-2007-moses}.
Non-verbal speech labels such as ``\textit{(Applause)}'' and  ``\textit{(Laughter)}'' were removed during evaluation~\citep{di-gangi-etal-2019-must,inaguma2020orthros,le-etal-2020-dual}.
We built output vocabularies based on the byte pair encoding (BPE) algorithm~\citep{sennrich-etal-2016-neural} with the Sentencepiece toolkit~\citep{kudo-2018-subword}\footnote{\url{https://github.com/google/sentencepiece}}.
The joint source and target vocabularies were constructed in the ST and MT tasks, while the vocabularies in the ASR task were constructed with transcriptions only.
For autoregressive models, we used $5$k for ASR models and $8$k for E2E-ST and MT models.
We used $16$k vocabularies for non-autoregressive E2E-ST models~\citep{inaguma2020orthros}.

For input speech features, we extracted $80$-channel log-mel filterbank coefficients computed with a $25$-ms window size and shifted every $10$ms with $3$-dimensional pitch features using Kaldi~\citep{kaldi}.
This results in $83$-dimensional features for every frame.
The features were normalized by the mean and the standard deviation for each training set.
To avoid overfitting, training data was augmented by a factor of $3$ with speed perturbation~\citep{speed_perturbation} and SpecAugment~\citep{specaugment}.
We used $(m_{T}, m_{F}, T, F)=(2, 2, 40, 30)$ for the hyperparameters in SpecAugment.

\subsection{Model configuration}\label{appendix:sec:model_configuration}
We used the Transformer~\citep{vaswani2017attention} architecture implemented with the ESPnet-ST toolkit~\citep{inaguma-etal-2020-espnet} for all tasks.
ASR and E2E-ST models consisted of $12$ speech encoder blocks and six decoder blocks.
The speech encoders had two CNN blocks with a kernel size of $3$ and a channel size of $256$ before the first Transformer encoder layer, which resulted in $4$-fold downsampling in the time and frequency axes.
The text encoder in the MT models consisted of six Transformer blocks.
The dimensions of the self-attention layer $d_{\rm model}$ and feed-forward network $d_{\rm ff}$ were set to $256$ and $2048$, respectively, and the number of attention heads $H$ was set to $4$.
For a large Transformer model configuration, we increased $d_{\rm ff}$ from $256$ to $512$ and $H$ from $4$ to $8$.
For a Conformer model configuration, we set $d_{\rm model}=256$, $d_{\rm ff}=2048$, and $H=4$.
The kernel size of depthwise separable convolution was set to $15$.
None of the other training or decoding hyperparameters were modified.

\subsection{Initialization}\label{appendix:sec:initialization}
In addition to initializing the encoder parameters of the E2E-ST model by those of the pre-trained ASR model, the auxiliary AR decoder parameters of the NAR models were initialized by those of the corresponding pre-trained AR MT model~\citep{inaguma2020orthros}.
The other decoder parameters of both the AR and NAR models were initialized as in BERT~\citep{devlin-etal-2019-bert,ghazvininejad-etal-2019-mask,inaguma2020orthros}, where weight parameters were sampled from ${\mathcal N}(0, 0.02)$, biases were set to zero, and layer normalization parameters were set to $\beta=0$, $\gamma=1$.
Note that we did not use additional data for pre-training.

\subsection{Training}\label{appendix:sec:training}
The Adam optimizer~\citep{adam} with $\beta_{1}=0.9$, $\beta_{2}=0.98$, and $\epsilon=10^{-9}$ was used for training with a Noam learning rate schedule~\citep{vaswani2017attention}.
We used dropout and label smoothing~\citep{label_smoothing} with a probability of $0.1$ and $0.1$, respectively.
The other training configurations for all tasks are summarized in Table~\ref{tab:training_config}.
We removed utterances having more than $3000$ input speech frames or more than $400$ characters due to the GPU memory capacity.
The last five best checkpoints based on the validation performance were used for model averaging.

For the training of ASR models used for E2E-ST encoder pre-training, we removed case and punctuation information from transcriptions and then applied a joint CTC/Attention objective~\citep{hybrid_ctc_attention}.
However, we retained this information in the transcriptions and paraphrases used for training the E2E-ST and MT models.

\begin{table}[t]
    \centering
    \begingroup
    \scalebox{0.78}{
    \begin{tabular}{l|cccc}\toprule
    \multicolumn{1}{c}{\multirow{2}{*}{Configuration}} & \multicolumn{1}{c}{\multirow{2}{*}{ASR}} & \multicolumn{2}{c}{E2E-ST} & \multirow{2}{*}{MT} \\ \cmidrule{3-4}
    \multicolumn{1}{c}{} & \multicolumn{1}{c}{} & AR & NAR &  \\
    \midrule
     Warmup step & 25k & 25k & 25k & 8k \\
     Learning rate factor & 5.0 (2.0) & 2.5 & 5.0 & 1.0 \\
     Batch size $\times$ accum & 128 & 128 & 256 & 96 \\ 
     Epoch & 45 (30) & 30 & 50 & 100 \\
     Validation metric & Accuracy & BLEU & BLEU & BLEU \\
     \bottomrule
    \end{tabular}
    }
    \endgroup
    \vspace{-2mm}
    \caption{Summary of training configuration. Numbers inside parentheses correspond to Conformer.}\label{tab:training_config}
    \vspace{-2mm}
\end{table}

\subsection{Case study}\label{appendix:sec:case_study}
We present examples of generated paraphrases on the Must-C En-De training set in Table~\ref{tab:result_case_study}.
We observed that most paraphrases kept the original meaning while some words were simplified to alternatives having a similar meaning.
We also found that the first conjunction in an utterance was more likely to be omitted via paraphrasing.

\subsection{Mathematical formulation of complexity and faithfulness}\label{appendix:mathematical_formulation}
In this section, we mathematically formulate the corpus-level complexity and faithfulness given ${\cal D} \in \{\stpair, \fwdseqkdstpair, \bwdseqkdstpair, \bidirseqkdstpair\}$.
Our formulation follows~\citep{zhou2019understanding}, but we also consider the reverse language direction.

\paragraph{Conditional entropy (complexity)}
The corpus-level complexity of ${\cal D}$ in the forward language direction, ${\cal C}(\overrightarrow{{\cal D}})$, is defined as the conditional entropy ${\cal H}(\ytgt|\ysrc)$ normalized over all samples.
${\cal H}(\ytgt|\ysrc)$ is defined as
\fontsize{9.7pt}{0pt}\selectfont
\begin{flalign*}
& {\cal H}(\ytgt|\ysrc) & \\
&= - \sum_{i=1}^{I} p(\ytgt_{i}|\ysrc_{i}) \cdot \text{log} \ p(\ytgt_{i}|\ysrc_{i}) & \\
&\approx - \sum_{i=1}^{I} (\prod_{k=1}^{|\ytgt_{i}|} p(\ytgt_{i,k}|\ysrc_{i})) \cdot \sum_{k=1}^{|\ytgt_{i}|} \text{log}\ p(\ytgt_{i,k}|\ysrc_{i}) & \\
&\approx - \sum_{k=1}^{T^{\rm t}} \sum_{y^{\rm t}_{k} \in {\cal A}(\ysrc_{i})} p(y^{\rm t}_{k}|\text{Align}(y^{\rm t}_{k})) \cdot \text{log} \ p(y^{\rm t}_{k}|\text{Align}(y^{\rm t}_{k})) & \\
&= \sum_{k=1}^{T^{\rm s}} {\cal H}(y^{\rm t}|\ysrc_{i,k}), &
\end{flalign*}
\normalsize
where ${\cal A}$ is an external alignment model, and $T^{\rm s}$ and $T^{\rm t}$ are the source and target sequence lengths, respectively.
We make two assumptions: (1) conditional independence of target tokens given the source text sequence, and (2) the distribution of $p(y^{\rm t}|\ysrc)$ follows the alignment model ${\cal A}$.
Then, ${\cal C}(\overrightarrow{{\cal D}})$ is calculated as
\begin{eqnarray*}
{\cal C}(\overrightarrow{{\cal D}}) &=& \frac{1}{|{\cal V}^{\rm s}|} \sum_{y^{\rm s} \in {\cal V}^{\rm s}} {\cal H}(y^{\rm t}|y^{\rm s}),
\end{eqnarray*}
where ${\cal V}^{\rm s}$ is a set of all words in the source language.
Division by $|{\cal V}^{\rm s}|$ is important to normalize frequent source words.

The corpus-level complexity of ${\cal D}$ in the backward language direction, ${\cal C}(\overleftarrow{{\cal D}})$, is defined similarly as
\begin{eqnarray*}
{\cal C}(\overleftarrow{{\cal D}}) &=& \frac{1}{|{\cal V}^{\rm t}|} \sum_{y^{\rm t} \in {\cal V}^{\rm t}} {\cal H}(y^{\rm s}|y^{\rm t}),
\end{eqnarray*}
where ${\cal V}^{\rm t}$ is a set of all words in the target language.

\paragraph{Faithfulness}
Although the corpus-level conditional entropy can be used to evaluate the complexity of the training data, there are also trivial solutions to generate new data with smaller complexity when target translations are not adequate.
Faithfulness is a good measure to assess how close the distilled data distribution is to the real (original) data distribution.
The faithfulness of ${\cal D}$ in a forward language direction ${\cal F}(\overrightarrow{{\cal D}})$ and a backward language direction ${\cal F}(\overleftarrow{{\cal D}})$ is defined as the KL-divergence of the alignment distribution between the real dataset and a distilled dataset, as
\begin{flalign*}
{\cal F}(\overrightarrow{{\cal D}}) &= \frac{1}{|{\cal V}^{\rm s}|} \sum_{y^{\rm s} \in {\cal V}^{\rm s}} \sum_{y^{\rm t} \in {\cal V}^{\rm t}} p_{\rm r}(y^{\rm t}|y^{\rm s})\ \text{log} \frac{p_{\rm r}(y^{\rm t}|y^{\rm s})}{p_{\rm d}(y^{\rm t}|y^{\rm s})}, & \\
{\cal F}(\overleftarrow{{\cal D}}) &= \frac{1}{|{\cal V}^{\rm t}|} \sum_{y^{\rm t} \in {\cal V}^{\rm t}} \sum_{y^{\rm s} \in {\cal V}^{\rm s}} p_{\rm r}(y^{\rm s}|y^{\rm t})\ \text{log} \frac{p_{\rm r}(y^{\rm s}|y^{\rm t})}{p_{\rm d}(y^{\rm s}|y^{\rm t})}, & 
\end{flalign*}
where $p_{\rm r}$ and $p_{\rm d}$ are alignment distributions of the real and distilled data, respectively.
Therefore, when ${\cal D}=\stpair$, ${\cal F}(\overrightarrow{{\cal D}})={\cal F}(\overleftarrow{{\cal D}})=0$, and it was omitted in Table~\ref{tab:analysis_faithfulness}.

\end{document}